\documentclass[letterpaper]{article} 
\usepackage{aaai23}  
\usepackage{times}  
\usepackage{helvet}  
\usepackage{courier}  
\usepackage[hyphens]{url}  
\usepackage{graphicx} 
\usepackage{booktabs}
\usepackage{natbib}  
\usepackage{caption} 
\usepackage{algorithm}
\usepackage[noend]{algorithmic}
\usepackage{multirow}

\usepackage{amsfonts}
\usepackage{bbm}
\usepackage{paralist}

\usepackage{amsmath}
\usepackage{pifont}

\usepackage{xcolor}

\usepackage{newfloat}
\usepackage{listings}
\DeclareCaptionStyle{ruled}{labelfont=normalfont,labelsep=colon,strut=off} 
\floatstyle{ruled}
\newfloat{listing}{tb}{lst}{}
\floatname{listing}{Listing}
\title{Audit and Improve Robustness of Private Neural Networks on Encrypted Data}
\author{
    Jiaqi Xue\textsuperscript{\rm 1}, 
    Lei Xu\textsuperscript{\rm 2}, Lin Chen\textsuperscript{\rm 3}, Weidong Shi\textsuperscript{\rm 4}, Kaidi Xu\textsuperscript{\rm 5}, Qian Lou\textsuperscript{\rm 1}
}
\affiliations{
    \textsuperscript{\rm 1 }University of Central Florida,
    \textsuperscript{\rm 2 }Kent State University, 
    \textsuperscript{\rm 3 }Texas Tech University,
    \textsuperscript{\rm 4 }University of Houston,
    \textsuperscript{\rm 5 }Drexel University  
   jiaq\_xue@knights.ucf.edu, xuleimath@gmail.com, lin.chen@ttu.edu, wshi3@central.uh.edu, kx46@drexel.edu, qian.lou@ucf.edu\\
}

\usepackage{bibentry}

\begin{document}

\maketitle

\begin{abstract}
Performing neural network inference on encrypted data without decryption is one popular method to enable privacy-preserving neural networks (PNet) as a service. 
Compared with regular neural networks deployed for machine-learning-as-a-service, PNet requires additional encoding, e.g., quantized-precision numbers, and polynomial activation.
Encrypted input also introduces novel challenges such as adversarial robustness and security. 
To the best of our knowledge, we are the first to study questions including \textit{(i) Whether PNet is more robust against adversarial inputs than regular neural networks?}\textit{(ii) How to design a robust PNet given the encrypted input without decryption?} 
We propose \textit{PNet-Attack} to generate black-box adversarial examples that can successfully attack PNet in both target and untarget manners. 
The attack results show that PNet robustness against adversarial inputs needs to be improved.
This is not a trivial task because the PNet model owner does not have access to the plaintext of the input values, which prevents the application of existing detection and defense methods such as input tuning, model normalization, and adversarial training.
To tackle this challenge, we propose a new fast and accurate noise insertion method, called \textit{RPNet}, to design Robust and Private Neural Networks. 
Our comprehensive experiments show that \textit{PNet-Attack} reduces at least $2.5\times$ queries than prior works. We theoretically analyze our RPNet methods and demonstrate that RPNet can decrease $\sim 91.88\%$ attack success rate.
\end{abstract}

\section{Introduction}

Machine-learning-as-a-service(MLaaS) is a powerful method to provide clients with intelligent services and has been widely adopted for real-world applications, such as image classification/segmentation, voice recognition, drug discovery, fraud detection, and many others~\cite{Delphi:usenix2020}. 
However, applying MLaaS to applications that involve biomedical, health, financial, and other sensitive data needs to protect data privacy~\cite{CryptoNets:ICML2016, SecureML:2017SP, GAZELLE:USENIX18, Delphi:usenix2020, Lou:ICLR2021:safenet}. 
By leveraging various cryptographic primitives, e.g., fully homomorphic encryption (FHE), secret sharing (SS), and multi-party computation (MPC), MLaaS providers can perform neural network inference in a privacy-preserving manner.
Specifically, the service providers only receive encrypted data from clients and process the data without decryption.
The processing results are also returned to the clients in cipher-text form~\cite{GAZELLE:USENIX18, Delphi:usenix2020}.
This type of MLaaS is usually referred to as PNet.
Although PNet was criticized for high performance, it is now efficient enough for real-world applications~\cite{capeprivacy, DualityTechnologies,Inpher,Zama}.

It is well-known that regular neural networks are vulnerable to  adversarial example attacks~\cite{goodfellow2014explaining, elsayed2018adversarial}, i.e.,  imperceptible perturbations onto the inputs can mislead regular neural networks to output wrong predictions. These imperceptible adversarial perturbations for black-box MLaaS have two main sources:
\begin{inparaenum}[(i)]
    \item Natural noise corruption on images, e.g., potential defects of the sensors and potential noises/damages of the optical devices~\cite{fu2022patchfool};
    \item Adversarial searches/estimations by attackers~\cite{byun2021small,qin2021random}.
\end{inparaenum}
As a special type of neural network, PNet also faces such risks, which is important when a PNet is deployed for real-world applications.

\begin{figure}[t!]
  \centering
   \includegraphics[width=\linewidth]{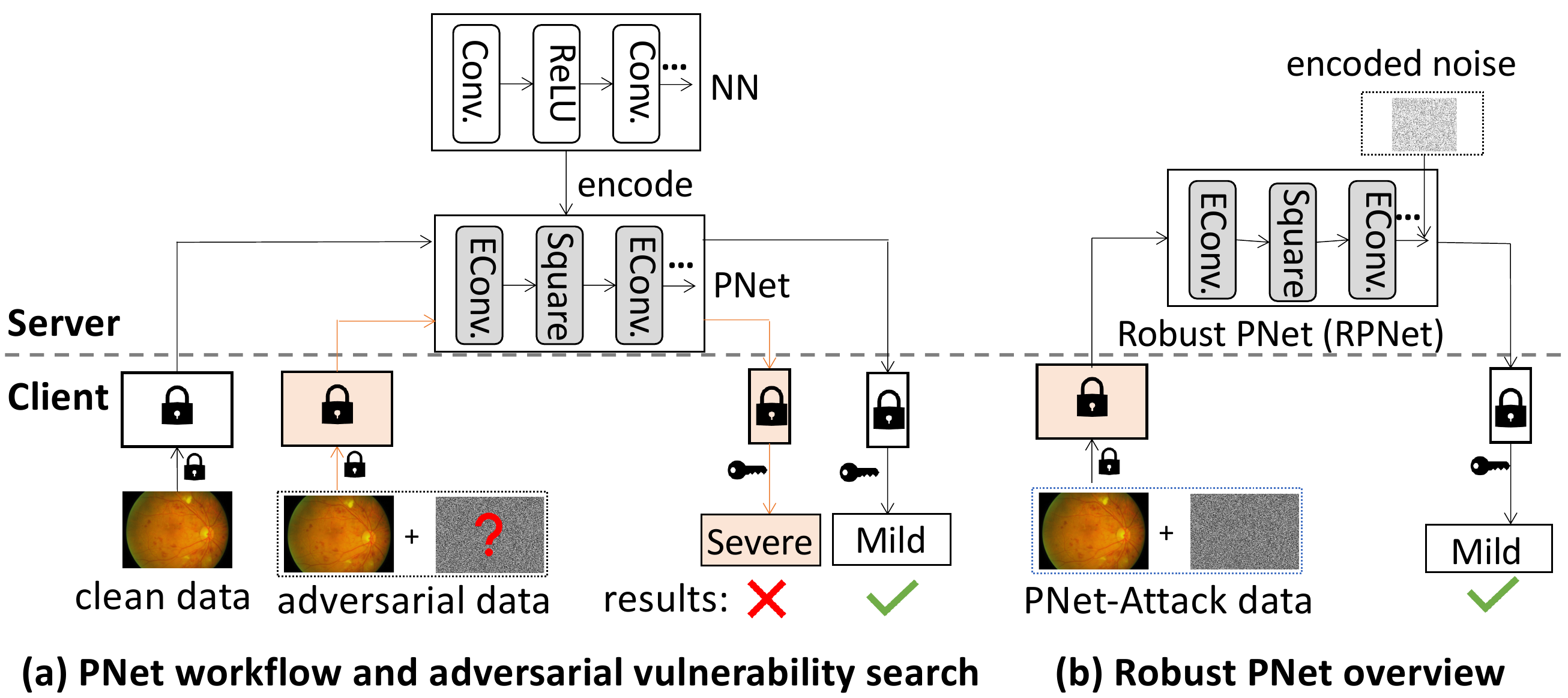}
   \caption{(a) Overview of the privacy-preserving neural network (PNet).   (b) Our RPNet enables a fast, accurate, and robust PNet against adversarial attacks by simply adding encoded noise to the output layer.}
   \label{fig:overview}
\end{figure}

It is not trivial to identify what potential natural noise corruptions and adversarial searches are vulnerable to PNet.
Our experiments show that PNet and regular neural networks have different adversarial robustness. Directly applying existing adversarial searches, e.g., SimBA~\cite{guo2019simple}, Square attack~\cite{andriushchenko2020square}, on PNet suffers from larger queries and lower attack success rate than regular neural networks. This is because PNet has a distinct workflow and features shown in Figure~\ref{fig:overview}(a), where the client submits encrypted data to the server that encodes a regular neural network (NN) into PNet to enable inference on encrypted data without decryption. The PNet encoding based on FHE involves two conversions, i.e., representing all the real values into integers or fixed-point numbers by quantization, approximating non-linear activation functions into approximated linear function, e.g., $square$ function. The inference result is also encrypted and only the data owner who has the private key can decrypt the result, thus this process is privacy-preserving. Nonlinear activation and highly quantized values induce PNet to have different robustness on the adversarial examples with the previous Net. 

Designing a robust PNet to defend the adversarial examples is also a great challenge. The main reason is that the defender, i.e., the server with the PNet model, takes encrypted data as input. It is imperceptible for the PNet owner whether the input contains adversarial perturbation, thus most of the existing defense methods dependent on input analysis are not applicable~\cite{NEURIPS2020_11f38f8e}. To defend against query-based black-box attacks, input-agnostic defense methods are related. Adding random noise into the input~\cite{qin2021random} or model~\cite{byun2021small} is a popular method to defend against attacks without perceiving the inputs. However, these methods are not designed for PNet and do not consider the distinct features of PNet, i.e., quantized activation and model, and polynomial activation, thus suffering from a very low defense success rate when adding a regular noise, or a large clean accuracy decrease when adding a larger noise.  

Our contributions can be summarized as follows. 
\begin{itemize}
    \item The robustness and security concerns of PNet against adversarial examples have not been studied. We first identified that directly applying existing attacks and defense techniques on PNet suffers from a low attack success rate and defense effects.
    
    \item We propose \textit{PNet-Attack} to efficiently attack PNet in both targeted and untargeted manners by an arc-shaped search in the frequency domain and a cosine annealing perturbation size schedule. 

    \item To defend the adversarial attacks, we propose \textit{RPNet} by adding noise in the output layer and a dynamic noise training (DNT) technique to design a Robust PNet.
    \item Our \textit{PNet-Attack} reduces $2.5\times \sim 3\times$ queries or increases $>18\%$ attack success rate than prior works. We theoretically analyze our \textit{RPNet} methods and our experiments demonstrate that \textit{RPNet} increases $>29\%$ and $>52\%$ targeted and untargeted attack failure rate with a $>0.36\%$ higher accuracy over prior defense works.
    
\end{itemize}

\begin{figure}[t!]
  \centering
   \includegraphics[width=0.9\linewidth]{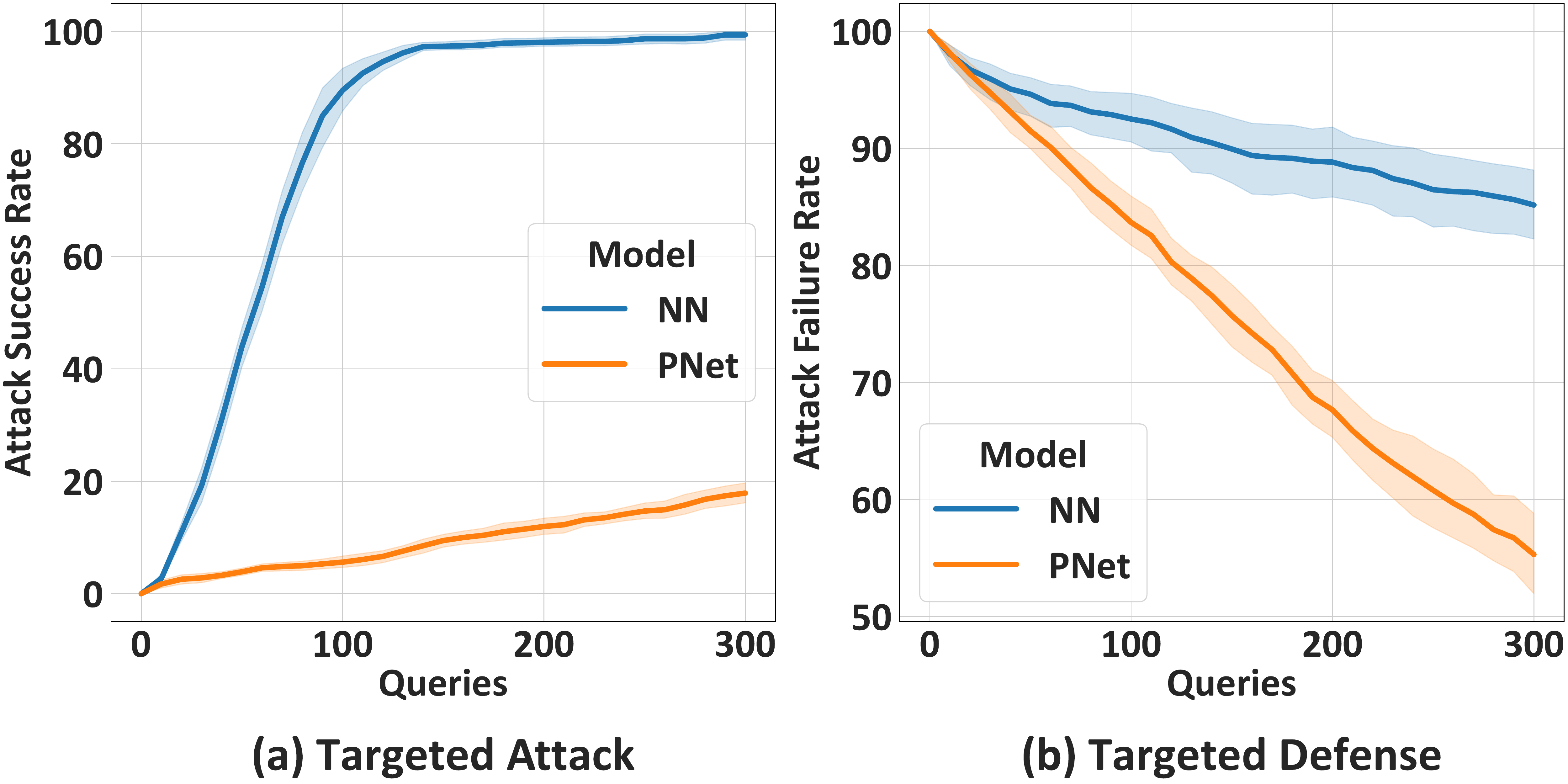}
   \caption{Existing black-box attacks and query-based defenses for regular NN are not transformed well to PNet.}
   \label{fig:motivation}
\end{figure}

\section{Background and Related Works}

\textbf{Privacy-Preserving Neural Network (PNet).}
Figure~\ref{fig:overview}(a) shows the workflow of the PNet, i.e., CryptoNets\cite{CryptoNets:ICML2016, FCryptoNets:arxiv19, Brutzkus:ICML19, Delphi:usenix2020}, where the client submits encrypted data to the server that encodes regular neural network (NN) into PNet to enable inference on encrypted data without decryption. The encoding converts real-number convolution (Conv.) into fixed-point one (Econv.), and replaces the nonlinear $ReLU$ function into polynomial functions, e.g., $square$ function. 
The inference result is also encrypted and only the data owner who has the private key can decrypt the result, thus this process is privacy-preserving. Nonlinear activation and highly quantized values induce PNet to have different robustness on the adversarial examples with the previous NN.

\noindent \textbf{Query-based Adversarial Attacks.}
For black-box MLaaS, adversarial examples can be from two sources: natural noise corruption on images, e.g., potential defects of the sensors or optical devices~\cite{fu2022patchfool} and query-based adversarial searches/estimations by attackers~\cite{byun2021small, qin2021random}. The key step of query-based adversarial attacks~\cite{attack1,attack2,attack4,attack5,attack6,attack8,attack9,attack10,attack11, guo2019simple} is to find an adversarial example perturbation direction to induce a decrease of designed objective by random search or gradient estimation during consecutive queries. Specifically, SimBA-DCT~\cite{guo2019simple} samples from an orthonormal bias and randomly searches the perturbations. Square attack~\cite{andriushchenko2020square} updates perturbations in a localized square-shaped and random area of the input. 
However, they are mainly evaluated in regular unencrypted neural networks and are not optimized for PNet. In contrast, our \textit{PNet-Attack} achieves more efficient attacks with fewer model queries by searching a frequency-domain adversarial perturbation in an arc-shaped order.

\noindent\textbf{Black-box Input-agnostic Defense.}
Since PNet takes encrypted data as input, the input content is imperceptible for the PNet owner, thus most of the existing defense methods dependent on input analysis are not applicable~\cite{NEURIPS2020_11f38f8e, chen2020stateful, li2020blacklight,pang2020advmind}. Input-agnostic defense methods have not been well studied but are required to defend against query-based black-box attacks. \cite{salman2020denoised, byun2021small} show that adding random noise into the input~\cite{qin2021random} or model~\cite{byun2021small} can defend against attacks without perceiving the inputs. Also, R\&P~\cite{xie2017mitigating} proposes an input random-transform defense method. RSE~\cite{liu2018towards} adds large Gaussian noise into both input and activation and uses ensembles to avoid accuracy decrease. PNI~\cite{he2019parametric, cohen2019certified, salman2019provably} incorporate noise in the training. However, these defense methods sacrifice enormous accuracy. And the input-transform function in R\&P and ensemble method in RSE introduce a large overhead for PNet. \cite{rusak2020simple} introduces that the model with Gaussian augmentation training could defend the common corruptions. RND~\cite{qin2021random} extends the methods in~\cite{rusak2020simple,byun2021small} and achieves the state-of-the-art defense against black-box attacks. However, RND does not consider the distinct features of PNet, i.e., quantized activation and model, and polynomial activation that has a decay effect on the added noise of the input, thus restricting the defense effect.  

\noindent\textbf{Threat Model.} We take one popular privacy-preserving cryptoNets~\cite{CryptoNets:ICML2016} as an example to present PNet, where the server hosts the PNet model and the clients submit encrypted data to request service. Our PNet can be easily extended to other hybrid private neural networks like Gazelle~\cite{GAZELLE:USENIX18} and Delphi~\cite{Delphi:usenix2020}. To fulfill the goal of deploying PNet in real-world MLaaS applications, the adversarial security and robustness concerns of PNet are also significant. These invisible adversarial examples have two main sources: natural noise corruption on images~\cite{fu2022patchfool} and adversarial searches/estimations~\cite{byun2021small,qin2021random}. 

Our PNet-attack shares the same threat model with Square attack~\cite{andriushchenko2020square} and SimBA~\cite{guo2019simple}, other than we use PNet-based MLaaS as the attack model shown in Figure~\ref{fig:overview}(a).  During each adversarial search, the adversarial example query is sent to the server after encryption. The server uses the encoded PNet to perform inference on encrypted data directly and returns the query result to the client. The client obtains the query result after decryption. Our Robust PNet (RPNet) shown in Figure~\ref{fig:overview} (b) follows the same threat model as RND~\cite{qin2021random} for a fair comparison. Also, we assume that the encrypted input is not accessible to the defender, i.e., the server, so the defense is black-box input-agnostic. 

\noindent\textbf{Limitations of Existing Attacks and Defenses on PNet.}
Existing black-box attacks and query-based defenses for Neural Network (NN) are not transformed well to PNet. Specifically, we use Figure~\ref{fig:motivation}(a) to show that one popular attack SimBA-DCT~\cite{guo2019simple} attains $\sim 80\%$ fewer attack success rates on PNet than NN for target attack.  This motivates us to design PNet-Attack to identify what adversarial examples are more vulnerable to PNet and how to generate them. Similarly, the encrypted input and additional encoding of PNet make the defense difficult. First, the encrypted input requires a black-box input agnostic defense which has not been well-studied. 
Second, the polynomial activation, i.e., degree-2 $square$ function, induces a decay effect on the added Gaussian noise of RND method especially when the absolute value of added noise is less than 1. We use  Figure~\ref{fig:motivation}(c) and (d) to show that compared to RND- defense in NN,  RND in PNet achieves  $\sim 32\%$ lower defense success rate (attack failure rate). This motivates us to design a robust PNet, RPNet, against adversarial attacks.    

\section{PNet-Attack}

In Figure~\ref{fig:overview}(a), we show that the adversarial example generation is a black-box query-based search. In particular, using the existing search method, e.g., SimBA-DCT~\cite{guo2019simple}, one can randomly update an adversarial example to decrease the designed objectives, e.g., misleading a prediction, during consecutive queries. However, existing methods have not been evaluated or optimized for PNet. Since PNet is performed on encrypted data, each query latency of PNet is 3 orders of magnitude higher than regular neural networks~\cite{CryptoNets:ICML2016, Lou:ICLR2021:safenet}. Therefore, designing a method that reduces the required query number for adversary example search is of great importance. 

To improve the searching efficiency, we propose a PNet-Attack method that is optimized for PNet shown in Algorithm~\ref{alg:attack}.
\begin{algorithm}[t!]
		\caption{PNet-Attack in Pseudocode }
		\label{alg:attack} 
		\begin{algorithmic}[1]
			\STATE $\textbf{Input:}$ image $x \in \mathbb{R}^{d\times d \times c}$, label $y$, step size seed $\epsilon$.
			\STATE adversarial perturbation $\delta = 0$
			\STATE $\mathbf{O} = M_p(x)$, $t=0$
            \STATE $\hat{x} = DCT(x)$   \;\;   \# for each channel
            \WHILE{$ \mathbf{O}_y = max_{y'}\mathbf{O}_{y'} \; and \;t< d^2 $}
            \STATE get $\hat{x_{i,j}}$ with the lowest frequency from  $\hat{x}$. 
            \STATE $\hat{x}=\hat{x}.pop(\hat{x_{i,j}})$ 
			\STATE$Q = Basis(\hat{x_{i,j}})$
             \FOR{$\hat{\alpha_t} \in \{\lambda_t \cdot \epsilon, -\lambda_t\cdot \epsilon \} $} 
              \STATE $t++$
              \STATE $\mathbf{O'} =  M_p(x+\delta_t + IDCT(\hat{\alpha_t} \cdot Q)) $
    		\IF{$sign(O^{'}_{y}-O_y)<0$} 
            \STATE $\delta_{t+1} = \delta_t + IDCT(\hat{\alpha_t} \cdot Q)$
            \STATE $\mathbf{O} = \mathbf{O'} $
            \STATE {$\textbf{break}$}
    		\ENDIF
  		    \ENDFOR
            \ENDWHILE	
		\bfseries{Return} {  $\delta$}
		\end{algorithmic}
	\end{algorithm}
The PNet-Attack method takes one clean image $X\in \mathbb{R}^{d\times d \times c}$, true label $y$, and step size seed $\epsilon$ as inputs, and generates adversarial perturbation $\delta$, where $d$ is the input width or height, $c$ is channel number. We define the prediction score probability of PNet model as $\mathbf{O}=M_p(x)$. Instead of adding perturbation in the spatial domain, we adopt a more efficient search direction $Q$ in the frequency domain by discrete cosine transform (DCT) and convert the frequency-domain perturbation $\hat{\alpha_t} \cdot Q$ back to the spatial domain by inverse DCT (IDCT). DCT and IDCT are defined in Appendix. The key idea of the algorithm is simple, i.e., for any direction $Q$ and step size $\hat{\alpha_t}$, one of $x+IDCT(\hat{\alpha_t} \cdot Q)$ or $x+IDCT(\hat{\alpha_t} \cdot Q)$ may decrease $\mathbf{O}=M_p(x)$. We iteratively pick direction basis $Q$ in the ascending order of frequency value $\hat{x_{i,j}}$ in $\hat{x}$. Note that we randomly sample one $\hat{x_{i,j}}$ when there are multiple entries with the equal value. For each query $t$, if the prediction probability $\mathbf{O'}$ of $x+IDCT(\hat{\alpha_t} \cdot Q)$ is decreased over $\mathbf{O}$, we will accumulate the perturbation $\delta_{t}$ with $IDCT(\hat{\alpha_t} \cdot Q)$, otherwise, we will subtract the $IDCT(\hat{\alpha_t} \cdot Q)$ from $\delta_{t}$.

The search efficiency of PNet-Attack algorithm is mainly dependent on two components, i.e., arc-shaped search order $Q$ in the frequency domain and perturbation size schedule $\lambda_t$.  In particular, frequency-domain input $\hat{x}$ is calculated by $DCT(x)$ for each channel, where the top-left positions of $\hat{x}$ have lower frequency values. Since low-frequency subspace adversarial directions have a much higher density than high-frequency directions, we try to perform the search from lower frequency to higher frequency before a successful attack. To achieve this goal, we iteratively extract the value $\hat{x_{i,j}}$ with the lowest frequency from $\hat{x}$. To avoid the repeating search, we pop out the $\hat{x_{i,j}}$ from the remaining search space $\hat{x}$ by $\hat{x} = \hat{x}.pop(x)$ shown in Algorithm~\ref{alg:attack}. The search direction basis $Q$ is set as $\hat{x_{i,j}}$ for the $t$-th query, which means that we only add the perturbation in the position of $\hat{x_{i,j}}$ and check if it decreases the prediction probability at the $t$-th query.  
\begin{equation} \small
    \lambda_t=  \lambda_{min} +\frac{1}{2}(\lambda_{max} - \lambda_{min})(1+cos(\frac{t}{T})\cdot \pi)
    \label{eq:schedule}
\end{equation}
Since $\hat{x_{i,j}}$ with lower frequency may contain more dense information than high-frequency values, we propose a perturbation size schedule $\lambda_t$ to assign larger perturbation size to the positions with lower frequency, which further improves the search efficiency. For $t$-th query, the perturbation size $\alpha_t$ is defined as the multiplication between $\lambda_t$ and frequency-domain perturbation seed $\epsilon$.  We define the cosine annealing schedule $\lambda_t$ in Equation~\ref{eq:schedule}, where $\lambda_{min}$ and $\lambda_{max}$ are the minimum and maximum coefficients of perturbation size, respectively, and  $\lambda_t\in [\lambda_{min}, \lambda_{max}]$, $T$ is the query range cycle. 
\begin{figure}[t!]
  \centering
   \includegraphics[width=\linewidth]{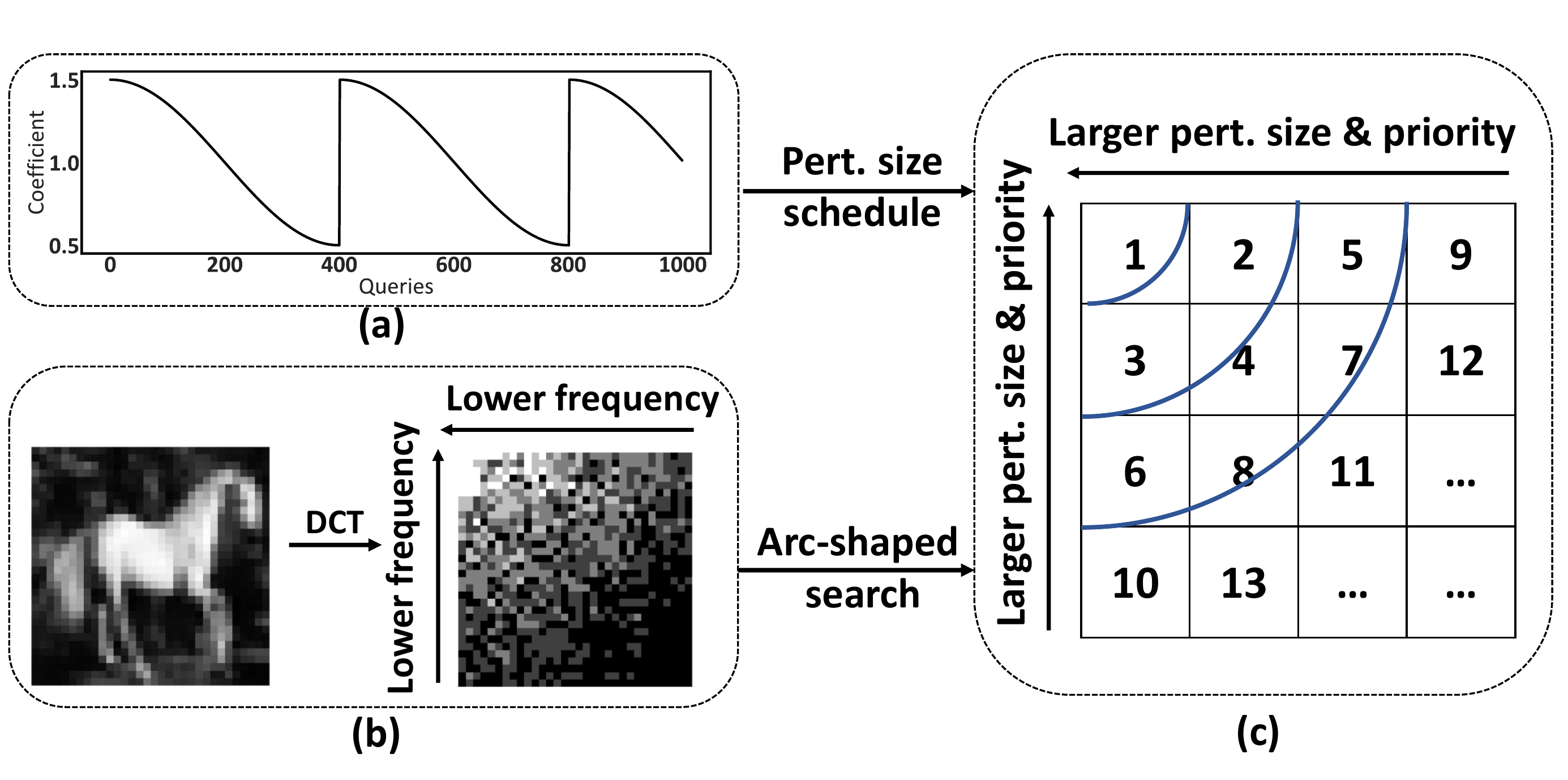}
   \caption{(a) Adversarial perturbation size over queries (b) frequency-domain input conversion by DCT. The top-left area has a lower frequency. (c) The perturbation order of PNet-Attack. PNet-attack assigns the position with a lower frequency to a larger perturbation size and priority. }
   \label{fig:attack_methods}
\end{figure}

In Figure~\ref{fig:attack_methods}, we illustrate the proposed PNet-Attack search algorithm and the workflow of generating an arch-shaped order and  perturbation size schedule. In particular, Figure~\ref{fig:attack_methods}(b) describes the conversion from input $x$ into frequency-domain $\hat{x}$ whose values decide the search priority in our Algorithm~\ref{alg:attack}, i.e., PNet-Attack. For example, the top-left position is of the highest priority to add perturbation, thus we perform the perturbation search on it in the first query. In Figure~\ref{fig:attack_methods}(c), we list the number of search orders, i.e., query orders. For instance, in the second query, we add perturbation on the element in the first-row second-column. This search order forms an arc shape, so we call this search order an arc-shaped search. Figure~\ref{fig:attack_methods}(a) demonstrates the perturbation schedule, i.e., the schedule size coefficient $\lambda_t$ over the query order. For example, the first query with the lowest frequency uses the largest $\lambda_t$ and the following queries with larger frequency will use smaller $\lambda_t$. 




\section{RPNet}

\textbf{Overview.} In query-based adversarial attacks, the attacker repeatedly adds a small perturbation to the input and checks if the consecutive two queries receive different prediction probabilities. If the objective prediction probability of $t+1$-th query is decreased over $t$-th query, the added perturbation is kept. Otherwise, the attacker subtracts the adversarial perturbation. The right perturbation search direction, i.e., adding or subtracting the perturbation in each query decides the search efficiency. Thus, a defender can disturb the perturbation search direction to decrease the attack efficiency by misleading the prediction probabilities. Motivated by those observations, we propose a fast and accurate defense method that simply adds noise to the output probability in each query to enable a robust and secure PNet, denoted by RPNet. Our RPNet defense method is designed to satisfy two objectives, i.e., the added defense noise will not significantly change the normal prediction probability so that the accuracy of clean data is not decreased, and the added defense noise will notably disturb the attack search direction and decrease the search efficiency. 

\begin{figure}[t!]
  \centering
   \includegraphics[width=\linewidth]{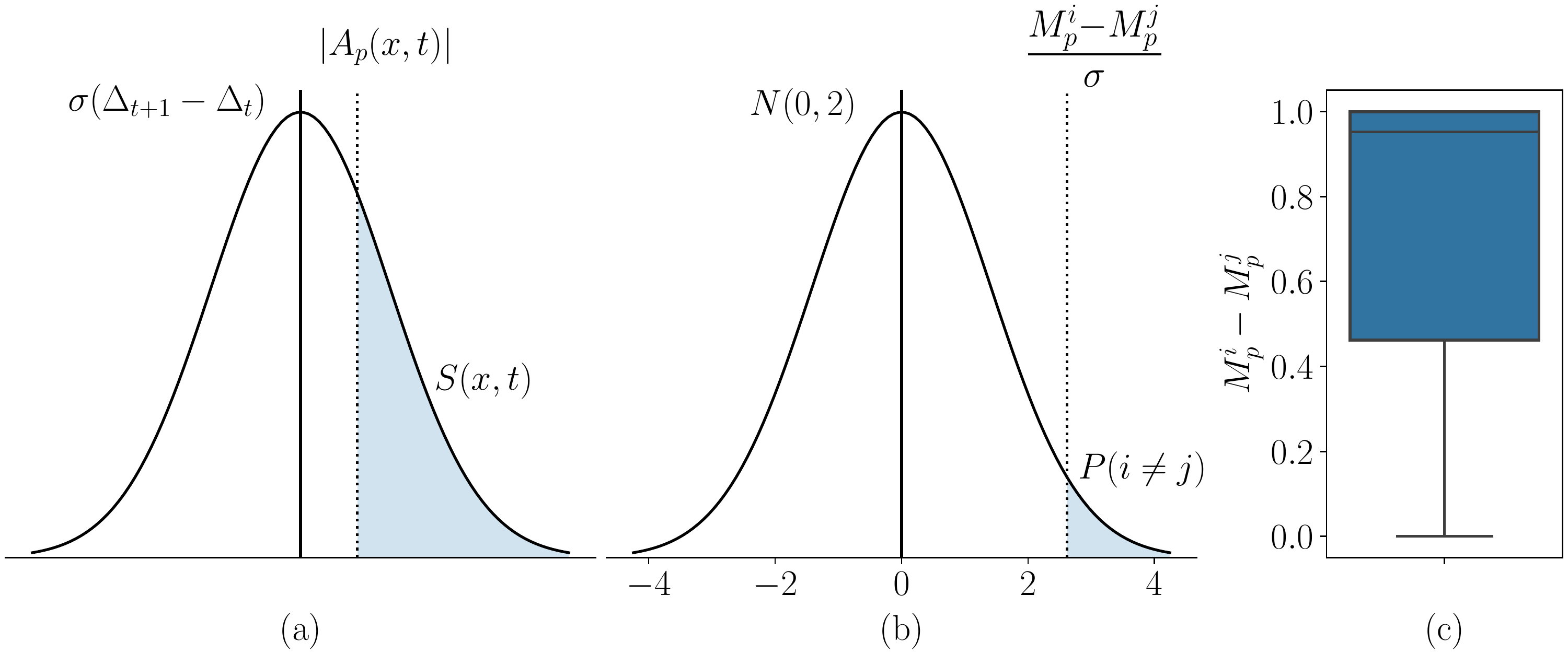}
   \caption{(a) disturbance success rate $S(x,t)$. (b) incorrect prediction rate $P(i\neq j)$. (c) $M_p^i-M_p^j$ distribution }
   \label{fig:probability}
\end{figure}

\noindent\textbf{RPNet Defense Formulation.}
We use Equation~\ref{eq:Apx} to define the prediction prabability difference of two queries on PNet, $M_p(x+\delta_{t}+\mu_t)$ and $M_p(x+\delta_t)$, where $\delta_t$ is the accumulated perturbation at $t$-th query, $\mu_t$ is the perturbation of $t$-th query, e.g., $IDCT(\hat{\alpha_t} \cdot Q)$ if defending PNet-Attack in Algorithm~\ref{alg:attack}. 
\begin{equation} \small
     \label{eq:Apx}
    A_p(x,t)=M_p(x+\delta_t+\mu_t)- M_p(x+\delta_t)
\end{equation}
In Equation~\ref{eq:Dpx},  we define the main step of the proposed RPNet defense method. One can see that two noises $\sigma \Delta_{t+1}$ and $\sigma \Delta_{t}$ are added into $(t+1)$-th query and $t$-th query, respectively, to disturb the attack search direction. Those noises are sampled from the same standard Gaussian distribution $\Delta \sim \mathcal{N}(0,1)$ and multiplied by a small factor $\sigma$. Note that the added noise shares the same encoding method with PNet for correct decryption of prediction result. The key idea of adding noise in the query result is to disturb the difference, i.e., $A_p(x,t)$, of two attack queries and mislead the search directions.  
\begin{equation} \small
\label{eq:Dpx}
\begin{aligned}
        D_p(x,t) &= (M_p(x+\delta_{t}+\mu_t)+ \sigma \Delta_{t+1})- (M_p(x+\delta_t) +\sigma \Delta_t) \\
        &=A_p(x,t)+\sigma (\Delta_{t+1}-\Delta _t)
\end{aligned}
\end{equation}
Specifically, the disturbance success happens when the signs of $A_p(x,t)$ and $D_p(x,t)$ are different. We use Equation~\ref{eq:Sx} to define the probability of disturbance success rate (DSR). A higher DSR will induce a lower attack success rate (ASR). Therefore, it is of great importance to understand the factors impacting the DSR. 
\begin{equation} \small
\label{eq:Sx}
    S(x,t)= P(sign (A_p(x,t)) \neq sign(D_p(x,t)))
\end{equation}
We theoretically analyze and calculate the DSR in Equation~\ref{eq:SxValue}. According to Equation~\ref{eq:Dpx}, the only difference of $D_p(x,t)$ and $A_p(x,t)$ is $\sigma (\Delta_{t+1}-\Delta _t)$, thus $S(x,t)$ is equal to the probability of adding $\sigma (\Delta_{t+1}-\Delta _t)$ to change the sign of $A_p(x,t)$. Given the Gaussian distribution $\sigma(\Delta_{t+1}-\Delta _t)\sim \mathcal{N}(0,2\sigma^2)$, the $S(x,t)$ is equal to $1- \phi (|A_p(x,t)|,;\mu =0, 2\sigma^2)$, where $\phi ()$ is the cumulative distribution function (CDF) of Gaussian distribution. This is because if $A_p(x,t)<0$, the added noise sampled from  $\mathcal{N}(0,2\sigma^2)$ should be larger than $|A_p(x,t)|$ to change the sign of $A_p(x,t)$, thus its probability is $1- \phi (|A_p(x,t)|,;\mu =0, 2\sigma^2)$; otherwise, the added noise should smaller than $|A_p(x,t)|$ to change the sign of $A_p(x,t)$, thus the probability is also $1- \phi (|A_p(x,t)|,;\mu =0, 2\sigma^2)$. Therefore, using the CDF equation, one can calculate the DSR in Equation~\ref{eq:SxValue}, where $erf$ is Gauss error function. We demonstrate in Equation~\ref{eq:SxValue} that DSR is impacted by two factors, i.e., $|A_p(x,t)|$ and $\sigma$. DSR has a positive relationship with $\sigma$ but is negatively relative to $|A_p(x,t)|$. In Figure~\ref{fig:probability} (a), we use the shaded area to illustrate the probability of S(x,t).
\begin{equation} \small
\label{eq:SxValue}
\begin{aligned}
    S(x,t) &= 1- \phi (|A_p(x,t)|;\mu =0, 2\sigma^2 )\\
    &= 1-\frac{1}{2\sigma \sqrt{\pi}} \int_{-\infty}^{|A_p(x,t)|}exp(\frac{(|A_p(x,t)|-\mu)^2}{-4\sigma^2})d|A_p(x,t)| \\
    &=\frac{1}{2}-erf(\frac{|A_p(x,t)|}{2\sigma})
\end{aligned}
\end{equation}
We also theoretically analyze the effects of our RPNet defense method on clean accuracy. When applying our PNet on a $n$-class classification task, we can define prediction score as $\mathbf{O} = \{M_p^0, M_p^1, ..., M_p^{n-1}\}$ for clean data. Since our defense method adds Gaussian noise $\sigma \Delta_t$ to the  $\mathbf{O}$, we define the prediction score after our defense method as $\mathbf{O^{\sigma}} = \{M_p^0+\sigma \Delta_t^0, M_p^1+\sigma \Delta_t^1, ..., M_p^{n-1}+\sigma \Delta_t^{n-1}\}$. The classification results of $\mathbf{O}$ and $\mathbf{O^{\sigma}}$ are $i=argmax(O)$ and $j=argmax(O^{\sigma})$, respectively. Therefore, RPNet will predict an incorrect classification if $i\neq j$. We use Equation~\ref{eq:Pij} to describe the probability of $P(i \neq j)$ that is positively relative to $\sigma$ but negatively relative to $M_p^i -M_p^j$.  In Figure~\ref{fig:probability} (b), we use the shaded area to illustrate the probability of $P(i\neq j)$. Our RPNet achieves a tiny $P(i\neq j)$ and a large $S(x,t)$ given a small $\sigma$, therefore obtaining an accurate and robust PNet. Figure~\ref{fig:probability} (c) demonstrates distribution of $M_p^i -M_p^j$ and most of values are larger than $0.5$ on CIFAR-10. The $\frac{M_p^i -M_p^j}{\sigma} >5$ since the $\sigma$ value is $<0.1$. Those observations show that $P(i\neq j)$ is tiny since 1-$\phi(\frac{M_p^i -M_p^j}{\sigma} >5;0,\sigma^2=2)$ is near zero. 

\begin{equation} \small
\label{eq:Pij}
\begin{aligned}
    P(i\neq j) &= P ((M_p^i + \sigma\Delta_t^i)< (M_p^j + \sigma\Delta_t^j))\\
    &= P(\frac{M_p^i -M_p^j}{\sigma}<\Delta_t^j-\Delta_t^i)\\
    &=\frac{1}{2}-erf(\frac{M_p^i -M_p^j}{2\sigma})
\end{aligned}
\end{equation}

\noindent\textbf{RPNet with dynamic noise training (DNT).} By analyzing the disturbance success rate $S(x,t)$ in Equation~\ref{eq:SxValue} and clean accuracy decrease rate $P(i\neq j)$ after applying our RPNet defense method, we reveal that a larger $\sigma$ will improve the defense effect but also may decrease the clean accuracy. To further avoid the clean accuracy decrease, one can reduce the noise sensitivity of PNet model or enlarge the difference between $M_P^i$ and $M_P^j$ in Equation~\ref{eq:Pij}. Inspired by those observations, we additionally equip our RPNet with dynamic noise training, denoted by RPNet-DNT, to enable a better balance between clean accuracy and defense effects. We use Algorithm~\ref{alg:RPNet-DNT} to describe RPNet-DNT that adds dynamic epoch-wise Gaussian noise $\sigma_i\Delta_t$ during each training iteration. Our results in Table~\ref{t:results_defense} show RPNet-DNT attains higher clean accuracy over RPNet.

\begin{algorithm}[h!]
		\caption{RPNet with dynamic noise training (DNT)}
		\label{alg:RPNet-DNT} 
		\begin{algorithmic}[1]
			\STATE $\textbf{Input:}$ RPNet model $M_p$, training data ($x, y$).
			\STATE  $t= 0$
            \FOR{$i=1$ to $epochs$}
            \STATE Randomly sample $\sigma_i \in [0, \sigma_{max}]$
			\FOR{$j=1$ to $iterations$} 
            \STATE $pred \; = M_p(x+\sigma_i \Delta_t)$
	        \STATE $t++$
         	 \STATE minimize $loss(pred,y)$ 
           \STATE  update  $M_p$
		      \ENDFOR
  		    \ENDFOR
		\bfseries{Return} {  $M_p$}
		\end{algorithmic}
	\end{algorithm}

\section{Experimental Setup}

\textbf{Datasets and Models.}
Consistent with previous privacy-preserving neural networks~\cite{CryptoNets:ICML2016, costache2017faster}, we conduct our experiments on MNIST~\cite{MNIST}, CIFAR-10~\cite{krizhevsky2014cifar}, and one medical datasets, Diabetic Retinopathy~\cite{medical_dataset}. On MNIST, we use the network defined in CryptoNets~\cite{CryptoNets:ICML2016} that has one convolution block, one square activation, and two fully connected (FC) layers. For other datasets, we follow the network that has three convolution blocks and two FC layers. Each convolution block consists of a convolutional layer, an activation square function, and an average pooling layer. The networks on MNIST, CIFAR-10, and medical datasets are quantized into 8 bits, 10 bits, and 16 bits respectively. 

\noindent\textbf{Evaluation Metrics.}
Attack success rate (ASR): the ratio of successful attack images to the total number of evaluated images. A larger ASR means a better attack performance. 
Average queries: the average number of queries per sample in the evaluated images, which equals the total number of queries divided by the total number of evaluated images. A smaller average query number represents a more efficient attack.
Average $\ell_2$ norm: the sum of $\ell_2$ norm of each adversarial image divided by the total number of evaluated images. A smaller average $\ell_2$ norm means a smaller adversarial size. Attack failure rate (AFR): the ratio of unsuccessful attack images. AFR is also the defense success rate. A larger AFR represents a better defense.
Clean Accuracy (ACC): the accuracy of clean data. Disturbance success rate (DSR): the probability of successfully disturbing the attack search direction. 

\noindent \textbf{Methodologies Study.} For attack methods, we compare prior works including SimBA-DCT~\cite{guo2019simple} and Square attack~\cite{andriushchenko2020square} with our PNet-Attack  without a schedule and with a schedule in Figure~\ref{fig:attack-queries} and Table~\ref{t:results_attack}. For defense methods, we compare prior works RND and its variant RND-GF~\cite{qin2021random} with our techniques including RPNet, RPNet with input noise, and RPNet-DNT. RPNet simply adds noise in the output layer shown in Equation~\ref{eq:Dpx}. RPNet+Input noise means adding the noise with a different scaling factor in the input layer. RPNet-DNT further incorporates the DNT technique.

\begin{table*}[ht!]
    \centering
\footnotesize
\setlength{\tabcolsep}{4pt}
\begin{tabular}{lcccccccccccc}\toprule
\multirow{3}{*}{Schemes}    & \multicolumn{6}{c}{CIFAR-10}  &   \multicolumn{6}{c}{Diabetic Retinopathy} \\
\cmidrule(lr){2-7}\cmidrule(lr){8-13}

   & \multicolumn{2}{c}{Average Queries}  &      \multicolumn{2}{c}{Average $\ell_2$} & \multicolumn{2}{c}{Success Rate}   & \multicolumn{2}{c}{Average Queries}  &      \multicolumn{2}{c}{Average $\ell_2$} & \multicolumn{2}{c}{Success Rate} \\

   \cmidrule(lr){2-3}\cmidrule(lr){4-5}\cmidrule(lr){6-7}\cmidrule(lr){8-9}\cmidrule(lr){10-11}\cmidrule(lr){12-13}
 & Untar.  &  Target     &   Untar.    & Target   &  Untar. & Target   & Untar.  &  Target                        &   Untar.  & Target      &  Untar. & Target \\

 \midrule
Square  &$100.1$     & $301.6$             &$4.21$  &$5.44$     & $85.64\%$ & $71.56\%$  &$50.0$     & $99.1$             &$1.47$  &$1.18$ &  $64.32\%$  & $64.14\%$\\
SimBA-DCT&$101.6$     & $302.4$             &$2.86$  &$4.81$     & $78.28\%$ &  $73.98\%$   &$51.8$     & $101.0$  &$0.82$  &$0.92$ & $73.28\%$ & $51.49\%$\\
PNet-Attack &$103.4$  & $299.4$   &$2.79$  &$4.09$   & $81.33\%$ &  $81.48\%$  &$50.5$     & $102.5$  &$0.84$  &$0.87$ & $84.36\%$ & $63.28\%$ \\
+Schedule &$99.8$     & $201.5$      &$3.61$  &$4.87$ & {$\mathbf{94.38}\%$} & {$\mathbf{94.22}\%$} &$50.4$     & $98.5$  &$1.15$  &$1.31$ & $\mathbf{89.92}\%$ & $\mathbf{76.48}\%$\\
\bottomrule
\end{tabular}
\caption{The attack comparisons of PNet-attack and prior works, e.g, SimBA-DCT~\cite{guo2019simple} and Square attack~\cite{andriushchenko2020square} on CIFAR-10 and medical dataset. 
Untar., Target means untarget and target attacks, respectively.
}
\label{t:results_attack}
\end{table*}


\begin{table*}[t]
    \centering
\footnotesize
\setlength{\tabcolsep}{4pt}
\begin{tabular}{lcccccccccccc}\toprule
\multirow{3}{*}{Schemes}    & \multicolumn{6}{c}{CIFAR-10}  &   \multicolumn{6}{c}{Diabetic Retinopathy} \\
\cmidrule(lr){2-7}\cmidrule(lr){8-13}

 & \multicolumn{2}{c}{Clean Accuracy}  &      \multicolumn{2}{c}{Average queries} & \multicolumn{2}{c}{Failure Rate}                & \multicolumn{2}{c}{Clean Accuracy}  &      \multicolumn{2}{c}{Average queries} & \multicolumn{2}{c}{Failure Rate} \\
   \cmidrule(lr){2-3}\cmidrule(lr){4-5}\cmidrule(lr){6-7}\cmidrule(lr){8-9}\cmidrule(lr){10-11}\cmidrule(lr){12-13}
 
 & Untar.  &  Target     &   Untar.    & Target   &  Untar. & Target   & Untar.  &  Target                        &   Untar.  & Target      &  Untar. & Target \\

\midrule
RND      &\multicolumn{2}{c}{$72.86\%$}  &$199.8$  &$301.1$     & $2.03\%$ &   $39.22\%$ &\multicolumn{2}{c}{$66.81\%$}   &$48.5$  &$53.1$ & $15.00\%$ & $46.40\%$\\
RND-GF &\multicolumn{2}{c}{$73.71\%$}  &$204.2$  &$300.4$     & $10.39\%$ &  $56.33\%$  &\multicolumn{2}{c}{$67.73\%$}  &$50.2$  &$49.3$ & $10.08\%$ & $59.60\%$\\
RPNet &\multicolumn{2}{c}{$74.10\%$}   &$198.2$  &$299.1$   & $56.17\%$ &  $88.04\%$  &\multicolumn{2}{c}{$67.91\%$}  &$51.7$  &$50.1$ & $49.77\%$ & $74.14\%$ \\
+Input noise &\multicolumn{2}{c}{$73.53\%$}    &$202.7$  &$300.1$     & $49.69\%$ &  $83.28\%$   &\multicolumn{2}{c}{$65.82\%$}   &$49.3$  &$50.7$ & $36.17\%$ & $74.53\%$ \\
+DNT &\multicolumn{2}{c}{$\mathbf{74.55}\%$}           &$199.4$  &$299.7$     & $\mathbf{63.36}\%$ &  $\mathbf{91.88}\%$  &\multicolumn{2}{c}{$68.09\%$}  &$48.9$  &$52.0$ & $\mathbf{66.41}\%$ & $\mathbf{88.67}\%$ \\
\bottomrule
\end{tabular}
\caption{The defense comparisons of RPNet and prior works, e.g, RND~\cite{qin2021random} and RND-GF~\cite{qin2021random}, on CIFAR-10 and medical dataset.  
'+Input noise', and '+DNT' represent adding Gaussian noise into the input layer and using an additional DNT method, respectively, on RPNet. }
\label{t:results_defense}
\end{table*}
\noindent \textbf{Parameter Settings.}
 We set the maximum number of queries for a single evaluated image as 300/100 for the targeted/untargeted attacks, respectively. For PNet-Attack, the cycle of schedule $T$ is 400, $\epsilon$ is 1, and $\lambda_{min}$ is 0.5, $\lambda_{max}$ is 1.5. For RPNet, we set $\sigma$ as 0.1. The scaling factor of input noise is set as $0.05$. For the defense method, the $\sigma_{max}$ is set as 0.25.  More experimental settings are included in Appendix. The results of the MNIST are shown in Appendix.  

\subsection{Experimental Results}

\begin{figure}[b!]
  \centering
   \includegraphics[width=\linewidth]{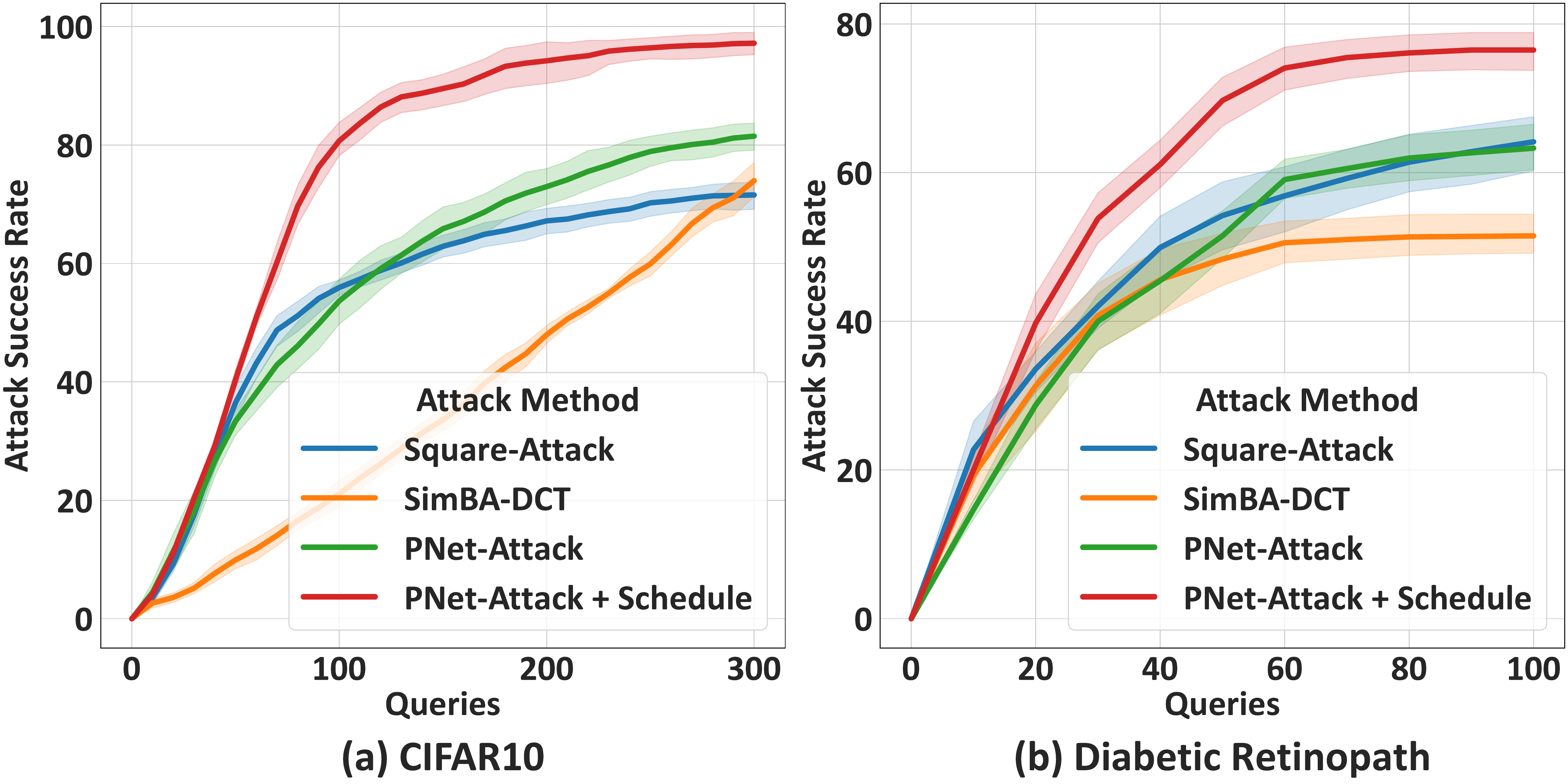}
   \caption{ Attack success rate v.s. number of queries for different methods on two datasets.
}
   \label{fig:attack-queries}
\end{figure}

\noindent\textbf{PNet-Attack Evaluation.}
In Table~\ref{t:results_attack}, we compare the attach performance of our PNet-Attack and previous works SimBA-DCT and Square attack on CIFAR-10 and medical dataset. For the targeted attack on CIFAR-10,  the previous method Square attack achieves 71.56\% attack success rate (ASR) with 301.6 average queries and 5.44 average $\ell_2$ norm. SimBA-DCT obtains 73.98\% ASR with an even smaller adversarial size, i.e., 4.81 $\ell_2$ norm. In contrast, our PNet-attack without a perturbation size schedule improves 7.5\% ASR with 0.72 smaller average $\ell_2$ perturbation norm over SimBA-DCT. Our perturbation size schedule further increases ASR of PNet-Attack by $12.74\%$ with $4.87$ average perturbation $\ell_2$ norm and only 201.5 average queries. Compared to Square attack, PNet-attack with schedule improves 22.66\% ASR, decreases 0.61 average $\ell_2$ norm, and reduces 100.1 queries. Compared to SimBA-DCT, PNet-attack with schedule increases 20.24\% ASR, decreasing 100.9 average queries with a similar average $\ell_2$ norm. Similarly, for the untargeted attacks,  PNet-attack with  schedule improves 8.74\% and 16.1\% ASR over Square attack and SimBA-DCT, respectively. The results of PNet-Attack on the medical Diabetic Retinopathy dataset also share a consistent trend with CIFAR-10.  

We show the attack processes of our PNet-Attack and previous works in Figure~\ref{fig:attack-queries}.  On both CIFAR-10 and Diabetic Retinopathy datasets, our PNet-Attack with schedule achieves higher targeted ASR than previous works SimBA-DCT and Square attack under the same queries. This is due to the fact that PNet-Attack with perturbation schedule and arc-shaped search order significantly improves the attack search efficiency. Our PNet-attack attains higher ASR than other techniques when using the same $\ell_2$ norm of adversarial example.  For example, PNet-Attack with schedule achieves $>20\%$ higher ASR than other techniques with $\sim 3.0$ average $\ell_2$ norm on the CIFAR-10 dataset.


\noindent\textbf{RPNet Defense Evaluation.}
In Table~\ref{t:results_defense}, we compare the defense effects of our RPNet and prior techniques, including RND and RND-GF proposed by~\cite{qin2021random}. For targeted attacks on CIFAR-10, RND realizes 39.22\% AFR with 72.86\% clean accuracy. RND-GF that adds Gaussian noise in the training accomplishes 56.33\% attack failure rate with 73.71\% accuracy. Different from RND which adds noise to the input, our RPNet adds noise to the confidence score, significantly improving AFR, i.e., $\sim 30\%$ AFR improvement. The reason is that adding noise into the input of PNet with polynomial activation will notably decay the noise, but adding the noise to the input can bypass the decay as demonstrated by our theoretical analysis of RPNet and experimental results. 
For instance, RPNet-DNT achieves 91.88\% AFR with even higher clean accuracy, 74.55\%. Compared to RND-GF, RPNet-DNT improves 35.55\% AFR and 0.84\% clean accuracy under the targeted attack on CIFAR-10. 

Our RPNet and RPNet-DNT have consistent improvements on the untargted attack and other medical datasets. In particular, RPNet-DNT attains 52.97\% untargeted attack failure rate improvement over RND-GF on CIFAR-10.  Similarly, on Diabetic Retinopathy dataset, RPNet increases 14.54\%, 39.69\% targeted and untargeted attack failure rates over RND-GF with 0.18\% higher clean accuracy. Adding noise into the input does not bring a significant improvement in attack failure rate, but the DNT technique remarkably brings a higher attack failure rate and clean accuracy. 

\begin{figure}[h!]
  \centering
   \includegraphics[width=0.9\linewidth]{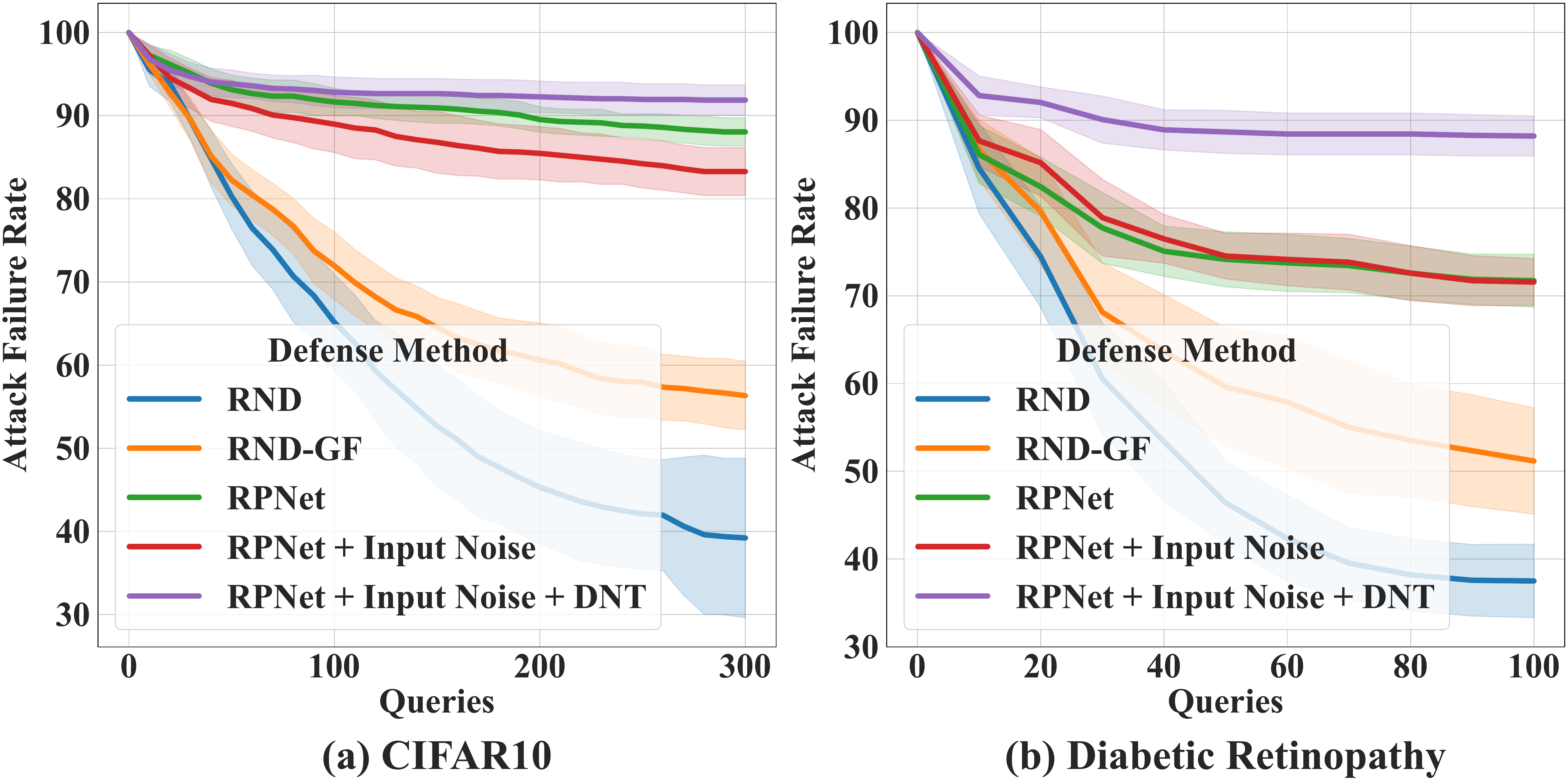}
   \caption{Defense effect comparisons of RPNet techniques and previous works. RPNet achieves a higher attack failure rate than prior work RND-GF. RPNet-DNT with input noise addition obtains the best performance.  }
   \label{fig:defense-queries}
\end{figure}

In Figure~\ref{fig:defense-queries}, we show the defense effects of prior works RND, RND-GF, and our techniques including RPNet, RPNet+Input noise, and RPNet-DNT with input noise. In particular, all techniques have a very high attack failure rate during the beginning queries since the attacks have a low attack success rate using few queries. With more queries, the attack failure rates of both RND and RND-GF are significantly reduced. In contrast, our RPNet still keeps a high attack failure rate. Similarly RND-GF, after adding noise into the input layer, the RPNet+Input noise does not bring a higher defense effect. This shows that the input noise added to the input may be decayed by the polynomial activation of PNet. With additional DNT techniques, RPNet-DNT further improves the defense effects. Note that without adding noise into the output layer, RPNet with input noise and DNT still cannot maintain a high defense effect. 

\begin{figure}[t!]
  \centering
   \includegraphics[width=0.9\linewidth]{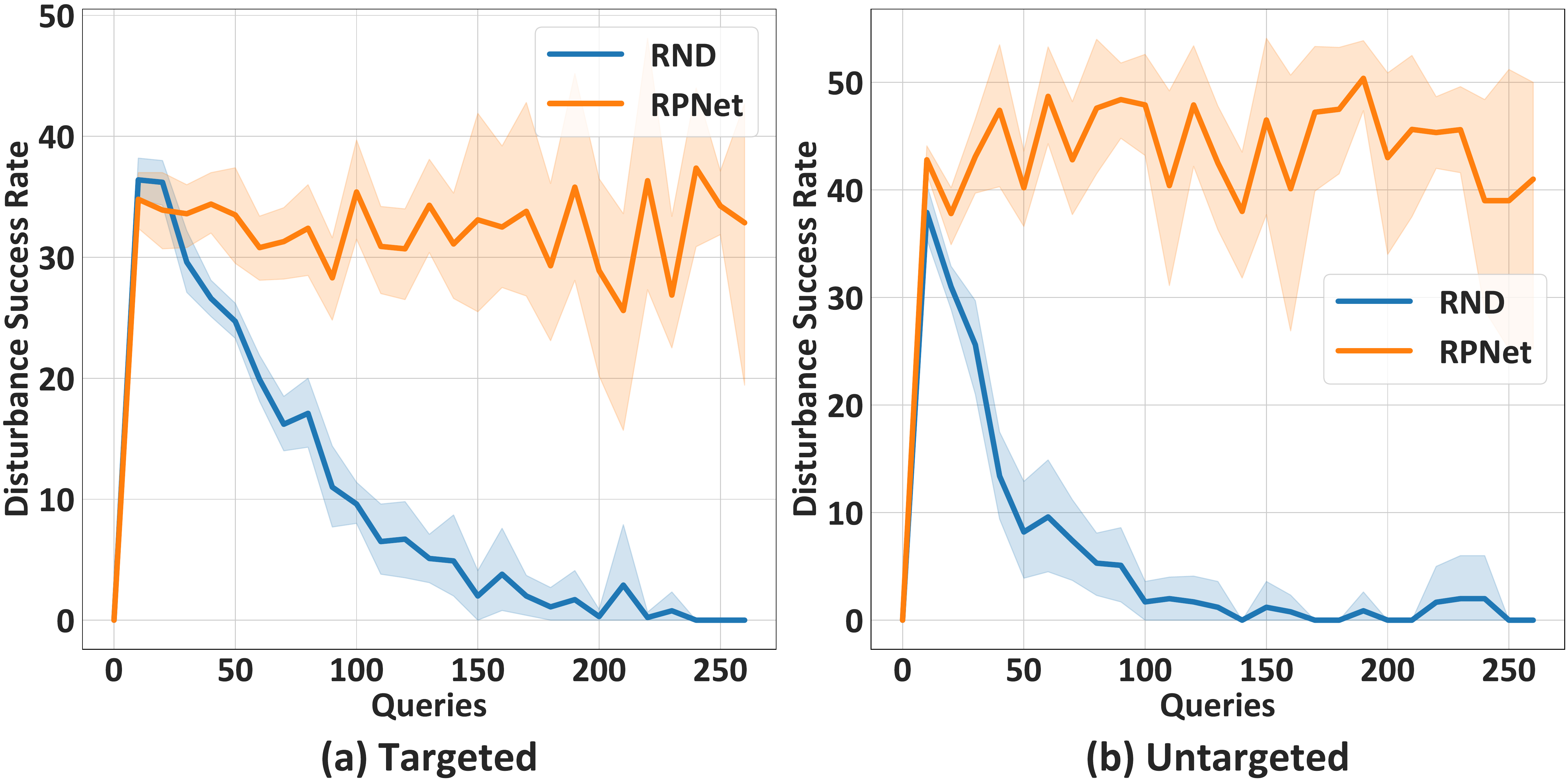}
   \caption{RPNet achieves a higher probability of disturbance success rate than prior work RND.}
   \label{fig:disturb-rate}
\end{figure}

We also use Figure~\ref{fig:disturb-rate} to show that RPNet has a larger defense efficiency than RND. 
RPNet attains a higher disturbance success rate than RND on both targeted and untargeted attacks, therefore empirically explaining the reason why RPNet achieves a higher defense effect.  We use Figure~\ref{fig:defense-sigma} in the appendix to show that RPNet achieves a better balance  of defense effect and accuracy than RND-GF. Specifically, Figure~\ref{fig:defense-sigma} (a)  illustrates that given a $\sigma$, e.g., 0.1, RPNet has a higher attack failure rate than RND-GF. Figure~\ref{fig:defense-sigma} (b) describes that give the same $\sigma$, e.g., 0.1, RPNet obtains a higher clean accuracy. RPNet has lower noise sensitivity than RND-GF.

\section{Conclusion}
We first identified that existing attacks and defense techniques for NN are not transformed well to PNet. We propose \textit{PNet-Attack} to efficiently attack PNet in both target and untarget manners by arc-shaped search in the frequency domain and a cosine annealing perturbation size schedule. To defend the adversarial attacks, we propose \textit{RPNet} by adding noise in the output layer and a DNT technique to design a Robust and Private Neural Network. 

\newpage
\bibliography{aaai23}

\bigskip
\noindent 
\clearpage
\section{Appendix}

\subsection{Ablation study}
\begin{figure}[h!]
  \centering
   \includegraphics[width=0.9\linewidth]{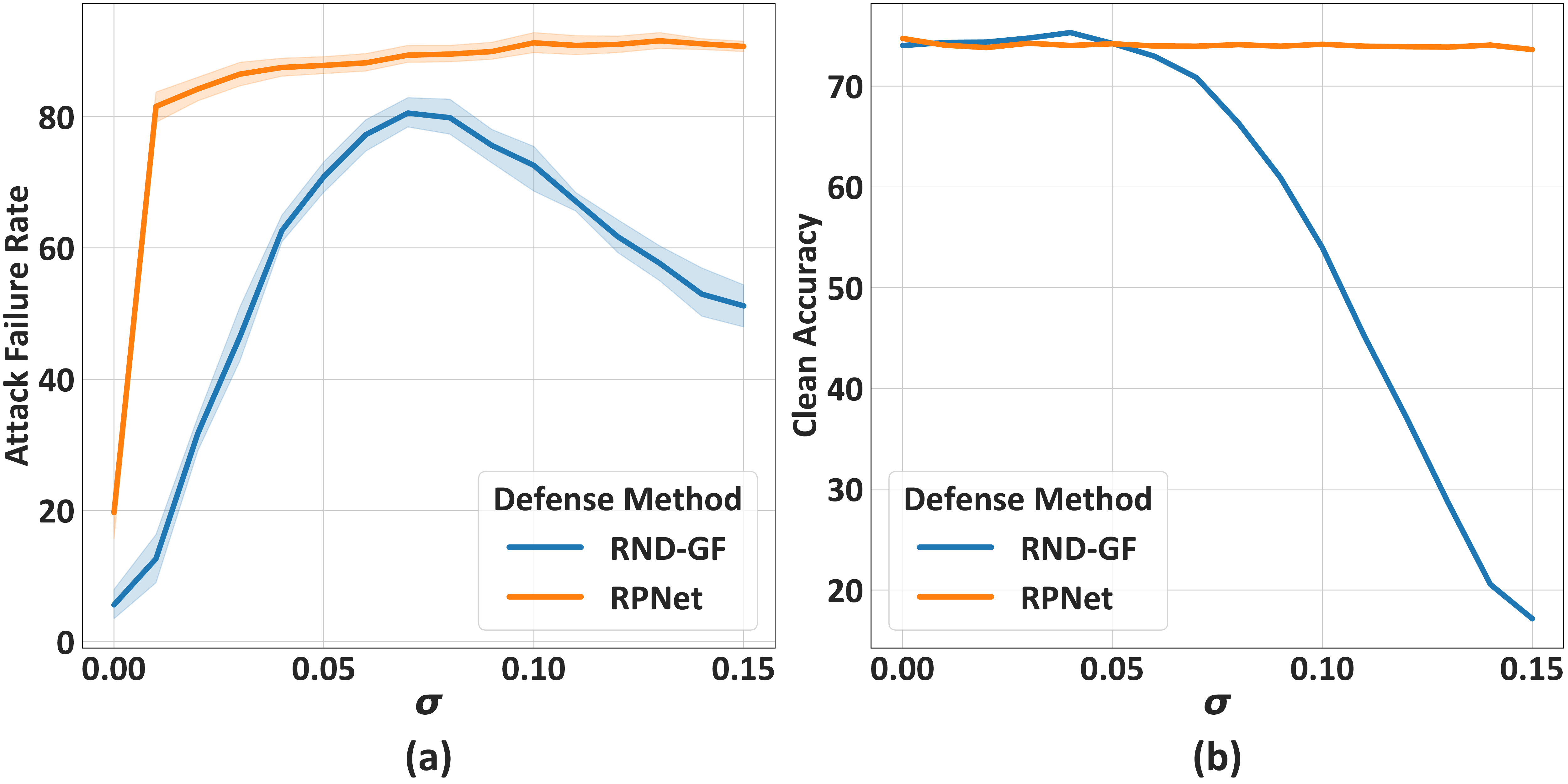}
   \caption{(a) Attack failure rate v.s. noise factor $\sigma$. (b) clean accuracy v.s. noise factor $\sigma$. RPNet achieves a better balance between defense effect and clean accuracy than prior work RND-GF.
   }
   \label{fig:defense-sigma}
\end{figure}

In Figure~\ref{fig:defense-sigma}, we show that RPNet achieves a better balance  of defense effect and accuracy than RND-GF. Specifically, Figure~\ref{fig:defense-sigma} (a)  illustrates that given a $\sigma$, e.g., 0.1, RPNet has a higher attack failure rate than RND-GF. Figure~\ref{fig:defense-sigma} (b) describes that give the same $\sigma$, e.g., 0.1, RPNet obtains a higher clean accuracy. RPNet has lower noise sensitivity than RND-GF.

\subsection{DCT and IDCT caculations}
The discrete cosine transform (DCT) represents an image as a sum of sinusoids of varying magnitudes and frequencies. Sepcifically, for an input image $X \in \mathbb{R}^{d\times d}$, the DCT transform $V=DCT(X)$ is:
\begin{equation} \small
    V_{m,n}=\alpha_m\alpha_n\sum_{i=0}^{d-1}\sum_{j=0}^{d-1}X_{i,j}\cos{\frac{\pi(2i+1)m}{2d}}\cos{\frac{\pi(2j+1)n}{2d}}
    \label{eq:schedule}
\end{equation}
where
\begin{equation} \small
    \alpha_m= \begin{cases}
    \sqrt{\frac{1}{d}},\quad &m=0 \\
    \sqrt{\frac{2}{d}},\quad &1\leq m\leq d-1
    \end{cases} 
    \label{eq:schedule}
\end{equation}
and
\begin{equation} \small
    \alpha_n= \begin{cases}
    \sqrt{\frac{1}{d}},\quad &n=0 \\
    \sqrt{\frac{2}{d}},\quad &1\leq n\leq d-1
    \end{cases} 
    \label{eq:schedule}
\end{equation}
for $0 \leq m,n \leq d-1$.

The values $V_{m,n}$ are called the DCT coefficients of. The DCT is an invertible transform, and its inverse IDCT is given by:
\begin{equation} \small
    X_{i,j}=\sum_{m=0}^{d-1}\sum_{n=0}^{d-1}\alpha_m\alpha_nV_{m,n}\cos{\frac{\pi(2i+1)m}{2d}}\cos{\frac{\pi(2j+1)n}{2d}}
    \label{eq:schedule}
\end{equation}
for $0 \leq i,j \leq d-1$. The basis functions are:
\begin{equation} \small
    \alpha_m\alpha_n\cos{\frac{\pi(2i+1)m}{2d}}\cos{\frac{\pi(2j+1)n}{2d}}
    \label{eq:schedule}
\end{equation}

\begin{figure}[h!]
  \centering
   \includegraphics[width=0.5\linewidth]{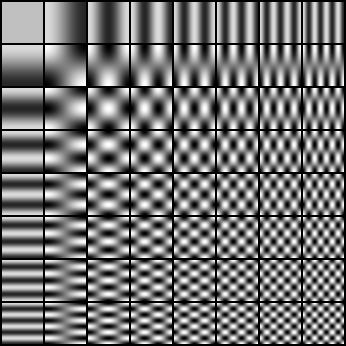}
   \caption{The 64 Basis Functions of an 8-by-8 Image.
   }
   \label{fig:DCT and IDCT}
\end{figure}

The IDCT equation can be interpreted as meaning that any d-by-d image can be written as a sum of basis functions. The DCT coefficients $V_{m,n}$ can be regarded as the weights applied to each basis function, with lower frequencies represented by lower $m,n$. Especially for 8-by-8 images, the 64 basis functions are illustrated by Figure~\ref{fig:DCT and IDCT}.
Horizontal frequencies increase from left to right, and vertical frequencies increase from top to bottom.

\subsection{Cryptosystems Settings}
For cryptosystems of PNet, one can follow the LoLa~\cite{Brutzkus:ICML19}, where the BFV scheme in SEAL~\cite{sealcrypto} is used. For MNIST and CIFAR-10, the plaintext modulus $m=2148728833\times 2148794369\times 2149810177$, modulus degree $N=16384$, coefficient modulus $q=\sim440$ bits. The security level is larger than 128 bits which is verified by $lwe\_estimator$~\cite{lwe_estimator}. To run PNets, one can run all experiments on the same Azure standard B8ms virtual machine with 8 vCPUs and 32GB DRAM.

\subsection{Results on MNIST dataset}
\begin{figure}[h!]
  \centering
   \includegraphics[width=0.9\linewidth]{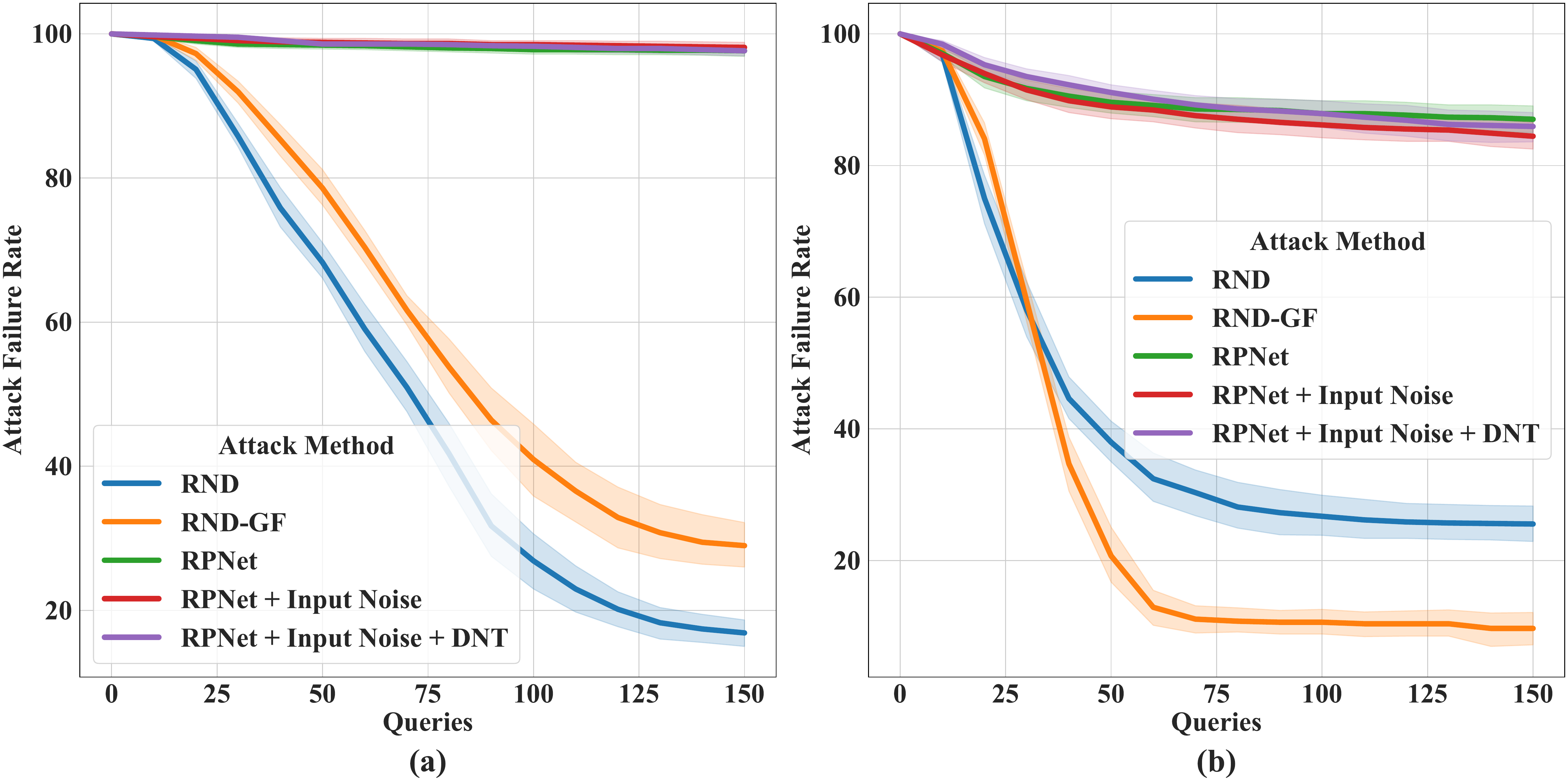}
   \caption{defense results on MNIST.
   }
   \label{fig:MNIST defense}
\end{figure}
In Figure~\ref{fig:MNIST defense}, we compare different defense methods on MNIST. We show that compared with the traditional methods, the AFR of RPNet proposed in our paper is greatly improved. 

For RND, $\sigma_1$=0.03. For RND-GF, $\sigma_1$=0.05. For RPNet, $\sigma$=0.05. For RPNet + Input Noise, $\sigma_1$=0.03, $\sigma$=0.05. For RPNet + Input Noise + DNT, $\sigma_1$=0.05, $\sigma$=0.05.
\end{document}